\documentclass[runningheads]{llncs}
\usepackage[T1]{fontenc}
\usepackage{graphicx}
\usepackage{multirow} 
\usepackage{makecell}
\usepackage{booktabs}
\usepackage{amsfonts}
\usepackage{ulem}
\usepackage{marvosym}

\def\etal{\normalem\emph{et al.~}}
\def\ie{\normalem\emph{i.e.}}
\def\eg{\normalem\emph{e.g.}}
\def\etc{\normalem\emph{etc.}}
\begin{document}
\title{Brain Lesion Synthesis via Progressive Adversarial Variational Auto-Encoder}

\titlerunning{Brain Lesion Synthesis via PAVAE}

\author{Jiayu Huo\inst{1}\textsuperscript{(\Letter)} \and
Vejay Vakharia\inst{2} \and
Chengyuan Wu\inst{3} \and
Ashwini Sharan\inst{3} \and
Andrew Ko\inst{4} \and
S\'{e}bastien Ourselin\inst{1} \and
Rachel Sparks\inst{1}
}

\authorrunning{J. Huo et al.}

\institute{School of Biomedical Engineering and Imaging Sciences (BMEIS), King's College London, London, UK \\
\email{jiayu.huo@kcl.ac.uk} \\
\and
National Hospital for Neurology and Neurosurgery, Queen Square, London, UK \\
\and
Division of Epilepsy and Neuromodulation Neurosurgery, Vickie and Jack Farber Institute for Neuroscience, Thomas Jefferson University, Philadelphia, Pennsylvania, USA \\
\and
Department of Neurosurgery, University of Washington, Seattle, Washington, USA
}


\maketitle              

\begin{abstract}
Laser interstitial thermal therapy (LITT) is a novel minimally invasive treatment that is used to ablate intracranial structures to treat mesial temporal lobe epilepsy (MTLE). Region of interest (ROI) segmentation before and after LITT would enable automated lesion quantification to objectively assess treatment efficacy. Deep learning techniques, such as convolutional neural networks (CNNs) are state-of-the-art solutions for ROI segmentation, but require large amounts of annotated data during the training. However, collecting large datasets from emerging treatments such as LITT is impractical. In this paper, we propose a progressive brain lesion synthesis framework (PAVAE) to expand both the quantity and diversity of the training dataset. Concretely, our framework consists of two sequential networks: a mask synthesis network and a mask-guided lesion synthesis network. To better employ extrinsic information to provide additional supervision during network training, we design a condition embedding block (CEB) and a mask embedding block (MEB) to encode inherent conditions of masks to the feature space. Finally, a segmentation network is trained using raw and synthetic lesion images to evaluate the effectiveness of the proposed framework. Experimental results show that our method can achieve realistic synthetic results and boost the performance of down-stream segmentation tasks above traditional data augmentation techniques.

\keywords{Laser interstitial thermal therapy \and 
Adversarial variational auto-encoder \and 
Progressive lesion synthesis.}
\end{abstract}

\section{Introduction}
Mesial temporal lobe epilepsy (MTLE) is one of the most common brain diseases and affects millions of people worldwide \cite{nevalainen2014epilepsy}. First-line treatment for MTLE is anti-seizure medicine but up to $30\%$ of patients do not achieve seizure control, in these patients resective neurosurgery may be curative \cite{rosenow2001presurgical}. As a minimally invasive treatment, laser interstitial thermal therapy (LITT) can accurately locate and ablate target lesion structures within the brain \cite{sun2015tissue}. LITT has been shown as an effective treatment for MTLE, and ablation of specific structures can be predictive of seizure freedom \cite{satzer2021extent}. Region of interest (ROI) segmentation needs to be performed to enable quantitative analyses of LITT \cite{vakharia2018automated} (\eg, lesion volume quantification and ablation volume estimation). However, manual delineation is inevitably time-consuming and requires domain knowledge expertise. Automated lesion segmentation could improve the speed and reliability of lesion segmentation for this task.


In the literature, some segmentation methods for the post-ablation area have already been exploited. Ermi{\c{s}} \etal \cite{ermics2020fully} developed Dense-UNet to segment resection cavities in glioblastoma multiforme patients. P{\'e}rez-Garc{\'\i}a \etal \cite{perez2021self} proposed an algorithm to simulate resections from preoperative MRI and utilized synthetic images to assist the brain resection cavity segmentation. Although the segmentation performance seems to be satisfied, it can be constrained by a small-scale dataset. Also, generated images of the rule-based resection simulation method can be less diverse, and imperfect synthetic results may compromise the performance of the segmentation model.

To mitigate the huge demand of images for training CNNs, methods utilizing generative adversarial network (GAN) \cite{goodfellow2014generative} have been presented. Han \etal \cite{han2018gan} generated 2D brain images with tumours from random noise to create more training samples. Kwon \etal \cite{kwon2019generation} implemented a 3D $\alpha$-GAN for brain MRI generation, including tumour and stroke lesion simulation. While these methods demonstrate the potential of GANs, there are some limitations. First, not all synthetic brain images have corresponding lesion masks, which means these methods may not be suitable to use for some down-stream tasks such as lesion segmentation. Additionally, these methods need extensive training samples to generate realistic results, which implies that the generalizability and robustness of these networks can not be ensured when the number of training samples is limited. Recently, Zhang \etal \cite{zhang2021carvemix} designed a lesion-ware data augmentation strategy to improve brain lesion segmentation performance. However, its effectiveness still can be affected due to limited training samples.

To address the aforementioned issues, we develop a novel progressive brain lesion synthesis framework based on an adversarial variational auto-encoder, and refer it as PAVAE. Instead of simulating lesions directly, we decompose this task into two smaller sub-tasks (\ie, mask generation and lesion synthesis) to alleviate the task difficulty. For mask generation, we utilize a 3D variational auto-encoder as the generator to avoid mode collapses. We adopt a WGAN \cite{arjovsky2017wasserstein} discriminator with the gradient penalty \cite{gulrajani2017improved} to encourage the generator to give more distinct results. We also design a condition embedding block (CEB) to encode semantic information (\ie, lesion size) to guide mask simulation. For lesion generation, we utilize a similar structure except replacing the CEB with a mask embedding block (MEB), to encode the shape information provided by masks to guide lesion synthesis. In the inference stage, we first sample from a Gaussian distribution to form the shape latent vector for mask simulation. Next, we combine the generated mask with an intensity latent vector sampled from a Gaussian distribution, and feed them into the lesion synthesis network to create the lesion image. Finally, we create new post-LITT ablation MR images from the generated lesion images. We train a lesion segmentation model using the framework nnUNet \cite{isensee2021nnu} to show the effectiveness of our method in synthesizing training images.


\section{Methodology}
Overall, the brain lesion synthesis task is decomposed into two smaller sub-tasks as described in Section \ref{section_model_architecture}. First, we design an adversarial variational auto-encoder to generate binary masks. To assist mask generation, we present a CEB to help encode mask conditions into the feature space, so that mask simulation can be guided by high-level semantic information. We adopt a similar architecture to generate lesions guided by binary lesion masks. Lesion masks are embedded into the feature space using a MEB. These additional blocks are described in Section \ref{section_condition_and_mask_embedding_block}. Finally, all models are trained using a four-term loss function as described in Section \ref{section_loss_functions} to ensure reconstructions are accurate, latent spaces are approximately Gaussian, and the real and simulated distributions are similar.

\subsection{Model Architecture}
\label{section_model_architecture}
\begin{figure}[!t]
\centering
\includegraphics[width=0.95\textwidth]{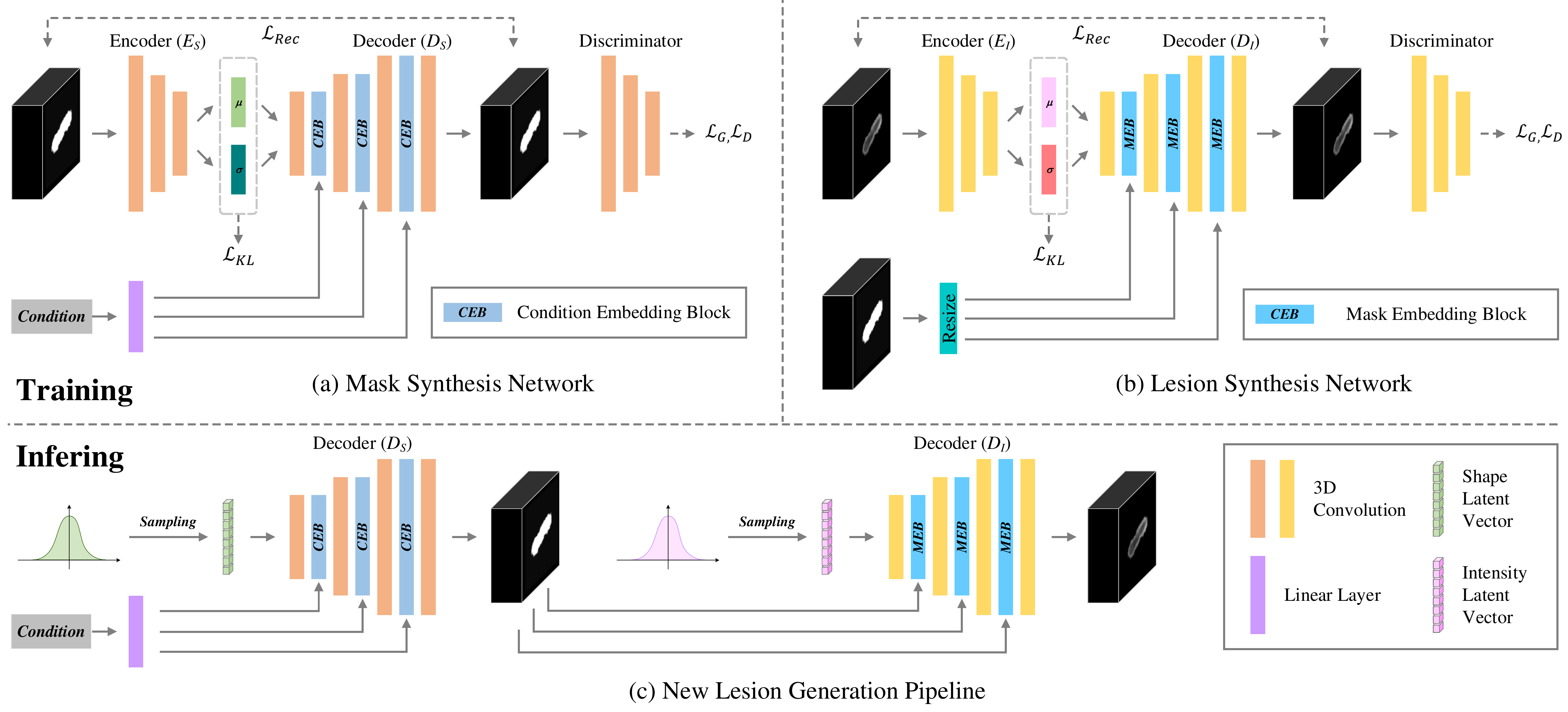}
\caption{The pipeline of our proposed framework for progressive brain lesion synthesis. Our method contains two separate networks with similar structures for (a) mask simulation and (b) lesion synthesis, respectively. For the inference (c), we sample shape latent vectors and intensity latent vectors from Gaussian distributions to generate new lesions.} 
\label{fig:framework}
\end{figure}

Fig. \ref{fig:framework} illustrates our progressive adversarial variational auto-encoder for brain lesion synthesis. We design a progressive 3D variational auto-encoder to approximate both shape and intensity information of post-LITT ablation lesions as Gaussian distributions. Besides, a following discriminator can ensure that generated images are more realistic. The kernel size of all convolutional layers is set to $3 \times 3 \times 3$, and Instance Normalization (InstanceNorm) and Leaky Rectified Linear Unit (LeakyReLU) are used after each convolutional layer. For the last convolutional layer, we use a Sigmoid function to scale output values between 0 and 1.

New lesion synthesis is performed as shown in Fig. \ref{fig:framework} (c). We first randomly sample from a Gaussian distribution to build shape latent vectors which are input into $D_{S}$ to generate new masks. Next, new masks and intensity latent vectors sampled from a Gaussian distribution are used as input for $D_{I}$ to generate new lesions. Here new masks are responsible for controlling new shapes and intensity latent vectors are responsible for intensity patterns.

\subsection{Condition and Mask Embedding Blocks}
\label{section_condition_and_mask_embedding_block}
\begin{figure}[!t]
\centering
\includegraphics[width=0.95\textwidth]{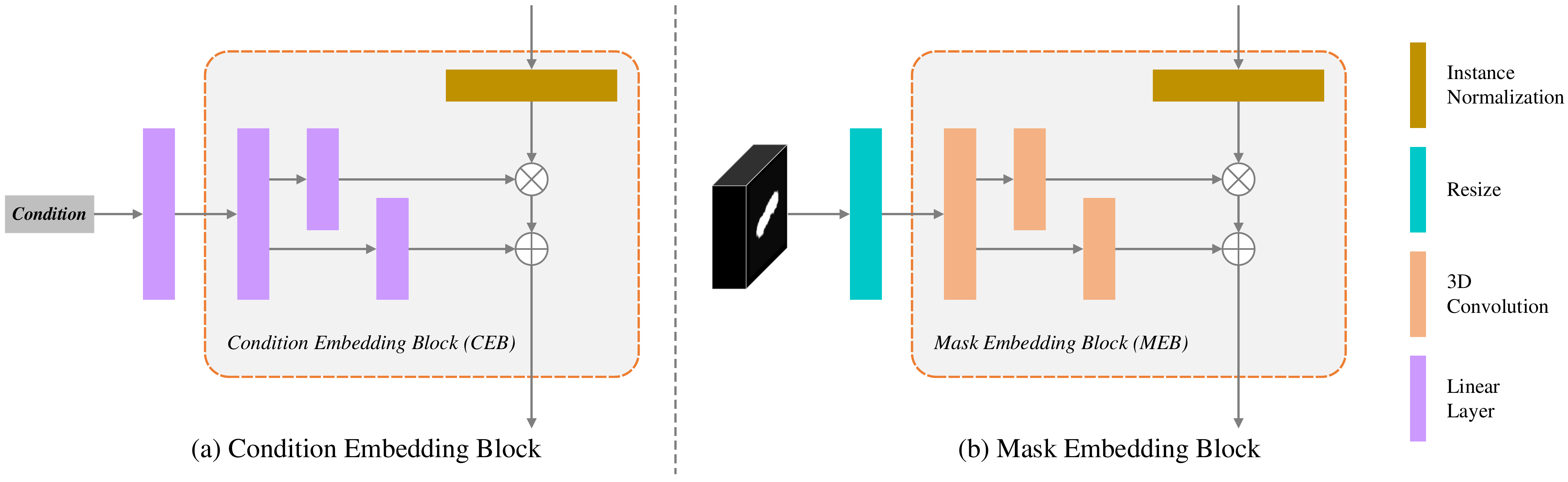}
\caption{The structure of (a) condition embedding block (CEB) and (b) mask embedding block (MEB).} 
\label{fig:ceb_and_meb}

\end{figure}

To add additional guidance for models in order to generate better results, we propose two separate modules shown in Fig. \ref{fig:ceb_and_meb}, a CEB and a MEB. For MEB, we follow the approach presented by SPADE \cite{park2019semantic}. First, masks are resized to the feature map resolution using nearest-neighbor downsampling. Next, learned scale and bias parameters are produced by three 3D convolutional layers. Finally, the normalized feature maps are modulated by the learned scale and bias parameters. For CEB, the structure is similar to MEB, but all 3D convolutional layers are replaced with linear layers since all input conditions are vectors.

\subsection{Loss Functions}
\label{section_loss_functions}
To optimize the encoder, decoder and discriminator so that reasonable masks and realistic lesions are generated, four loss functions are utilized in our work: reconstruction Loss $\mathcal{L}_{Rec}$, KL divergence $\mathcal{L}_{KL}$, and GAN specific losses $\mathcal{L}_{G}$ and $\mathcal{L}_{D}$. First, the reconstruction loss $\mathcal{L}_{Rec}$ is used to ensure outputs have high fidelity to the ground truth images. The Kullback-Leibler (KL) loss $\mathcal{L}_{KL}$ is imposed on the model to minimize the KL divergence between the intractable posterior distribution and the prior distribution (\ie, Gaussian distribution in latent space). Furthermore, we add Wasserstein loss functions ($\mathcal{L}_{G}$ and $\mathcal{L}_{D}$) to the 
GAN in order to prevent results generated by the decoder from being fuzzy.

The reconstruction loss implemented in our framework is mean squared error (MSE) defined as:
\begin{equation}
{\mathcal{L}_{Rec}} = \sum\nolimits_i {{{\left\| {x_g^{\left( i \right)} - x_r^{\left( i \right)}} \right\|}^2}},
\label{eq:loss_rec}
\end{equation}
where $x_r^{\left( i \right)}$ refers to the $i^{th}$ real image within a mini-batch, and $x_g^{\left( i \right)}$ denotes the $i^{th}$ generated image obtained from the decoder. MSE guarantees that real images and synthetic images look similar in general. However, synthetic images may lose some detailed information, which will make them appear indistinct.


The KL loss is defined as the KL divergence $\mathcal{D}_{KL}$ between the posterior distribution $q\left(z|\cdot \right)$ and the prior distribution $p\left(z\right)$, which is formulated as:
\begin{equation}
{\mathcal{L}_{KL}} = \sum\nolimits_i {{\mathcal{D}_{KL}}\left[ {q\left( {z|x_r^{\left( i \right)}} \right)\parallel p\left( z \right)} \right]},
\label{eq:loss_kl}
\end{equation}
where $q\left( {z|x_r^{\left( i \right)}} \right)$ is the posterior latent distribution under the condition of $x_r^{\left( i \right)}$, and $p\left(z\right)$ is a normal distribution for the latent vector $z$. By minimizing the KL divergence between the two distributions, the conditional distribution of the latent vector $z$ approximates a Gaussian distribution.


To avoid generating images with blurriness and instability during training, we deploy the loss functions from WGAN \cite{arjovsky2017wasserstein} instead of the original GAN. The Wasserstein loss can be defined as:
\begin{equation}
{\mathcal{L}_{D}} = \mathbb{E}_{x_{g} \sim \mathbb{P}_{g}}\left[D\left(x_{g}\right)\right]-\mathbb{E}_{x_{r} \sim \mathbb{P}_{r}}\left[D\left(x_{r}\right)\right]+\lambda \mathbb{E}_{\hat{x} \sim \mathbb{P}_{\hat{x}}}\left[\left(\left\|\nabla_{\hat{x}} D(\hat{x})\right\|_{2}-1\right)^{2}\right],
\label{eq:loss_d}
\end{equation}
\begin{equation}
{\mathcal{L}_{G}} = - \mathbb{E}_{x_{g} \sim \mathbb{P}_{g}}\left[D\left(x_{g}\right)\right].
\label{eq:loss_g}
\end{equation}
Compared with original GAN using a discriminator to differentiate whether images are real or fake, WGAN uses the Wasserstein distance to directly estimate the difference between two distributions $\mathbb{P}_{r}$ and $\mathbb{P}_{g}$. This Wasserstein distance can be formulated as $\mathcal{D}_{W} = \mathbb{E}_{x_{r} \sim \mathbb{P}_{r}}\left[D\left(x_{r}\right)\right]-\mathbb{E}_{x_{g} \sim \mathbb{P}_{g}}\left[D\left(x_{g}\right)\right]$, where $\mathbb{E}$ denotes the maximum likelihood estimation and $D(\cdot)$ denotes the discriminator. In addition, we further include a gradient penalty regularization \cite{gulrajani2017improved} to constrain $\mathcal{L}_{D}$ to satisfy the 1-Lipschitz condition, so that $\mathcal{L}_{D}$ will remain stable during network training. The gradient penalty regularization is formalized as $\lambda \mathbb{E}_{\hat{x} \sim \mathbb{P}_{\hat{x}}}\left[\left(\left\|\nabla_{\hat{x}} D(\hat{x})\right\|_{2}-1\right)^{2}\right]$, where $\hat{x}$ denotes random interpolation between real samples and generated samples, and $\lambda$ is a weighting factor. In our experiments, we fix the value of $\lambda$ to 10.

\section{Experiments}

\subsection{Dataset}
In this study, 47 T1-weighted MRI scans of 47 patients are collected from a high-volume epilepsy surgery center which has already established expertise in using LITT for MTLE. Consecutive patients are included if they received LITT for MTLE and had concordant semiology, scalp electroencephalography (EEG) and structural MRI features of mesial temporal sclerosis, or had seizure onset confirmed within the hippocampus following stereo-EEG (SEEG) investigation. Ethical approval for the study was provided by institutional review board approval for the retrospective use of anonymized imaging. All T1-weighted images are first aligned to the MNI152 brain template \cite{Fonov2009UnbiasedNA}. A random split of the dataset is performed, keeping $20\%$ (10 cases) of the whole dataset as the test set with the remaining $80\%$ (37 cases) being used as the training set.

\subsection{Implementation Details}
Our framework is implemented within PyTorch 1.10.0 \cite{paszke2019pytorch}. For network training, encoder and decoder layers are treated as the generator and optimized together. To optimize the generator and discriminator networks, we use two Adam optimizers \cite{kingma2014adam}. The initial learning rate is set to 5$e$-5, and the batch size is set to $13$ due to GPU memory limitations. For each model, we train for $1000$ epochs individually using only data in the training set. For input images, we extract a $64 \times 64 \times 64$ cube from raw MRI scans corresponding to a ROI containing the mask to ensure the entire lesion area is included within the image.

\subsection{Evaluation Metrics}
All metrics are evaluated and reported on the test set. First, we evaluate the lesion synthesis performance using three metrics, \ie, peak signal-to-noise ratio (PSNR), structural similarity (SSIM), and normalized mean square error (NMSE). Moreover, to prove that generated brain lesions can further boost the performance of down-stream tasks, we use four metrics to measure brain lesion segmentation performance: Dice coefficient, Jaccard index, the average surface distance (ASD), and the $95\%$ Hausdorff Distance (95HD).

\subsection{Experimental Results}

\subsubsection{Comparison of Lesion Synthesis Performance}
\begin{figure}[!t]
\centering
\includegraphics[width=0.95\textwidth]{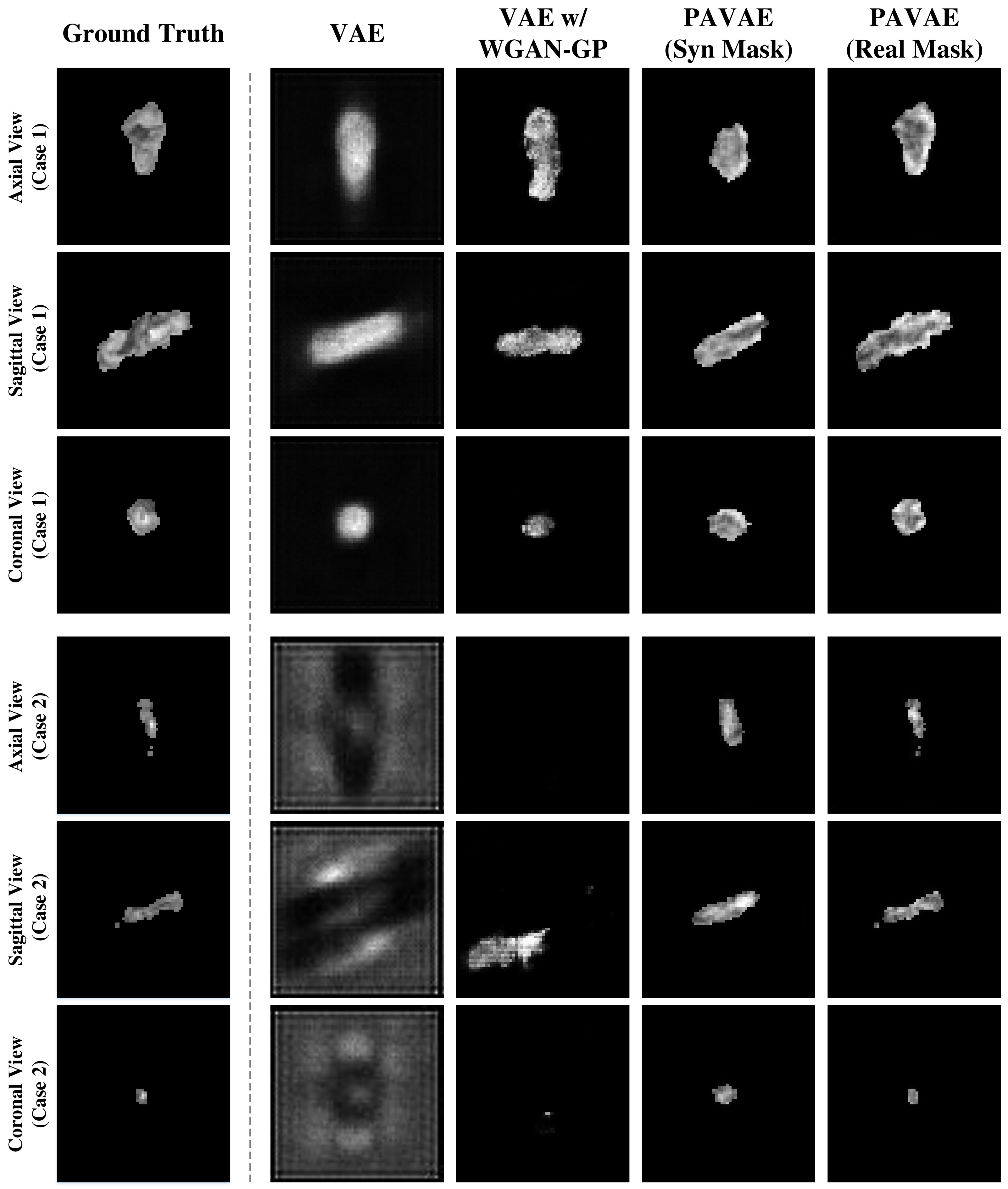}
\caption{Qualitative comparison for different generative models.} 
\label{fig:syn_viz}
\end{figure}

We first qualitatively compare the synthetic results of our framework with other existing methods. Specifically, we employ 3D VAE and 3D VAE w/ WGAN-GP as baseline models. For 3D VAE, only $\mathcal{L}_{Rec}$ and $\mathcal{L}_{KL}$ is utilized for model training. For 3D VAE w/ WGAN-GP, the structure is similar to the lesion synthesis network and all loss functions are utilized for training, but MEB has not been included. Note that all models are implemented with 3D convolution and 3D InstanceNorm layers. Triplanar views of synthetic lesions are shown in Fig. \ref{fig:syn_viz}. Here, 3D VAE w/ WGAN-GP refers to 3D VAE followed by a WGAN-GP discriminator, PAVAE (Syn Mask) indicates lesion synthesis utilizes generated masks derived from the mask synthesis network. PAVAE (Real Mask) indicates that real masks are utilized to guide lesion synthesis. As can be found in Fig. \ref{fig:syn_viz}, 3D VAE always generates fuzzy lesion images. Also, small lesions seem to be diffused, indicating the model has trouble in simulating small lesions. When a WGAN-GP discriminator is added to 3D VAE, results are clearer for big lesions. However, WGAN-GP still can not simulate small lesions. For our model, using synthetic masks to guide lesion generation \ie, PAVAE (Syn Mask), we observe even small lesions are successfully generated. Utilizing real masks for guidance results in synthetic lesions are closest to the ground truth among all compared methods (the rightmost column in Fig. \ref{fig:syn_viz}). This highlights that the lesion synthesis network can generate realistic image intensity when provided a realistic lesion mask.

\begin{table}[!t]
\caption{Quantitative comparison of synthetic results for different generative models.}
\label{tab:syn_results}
\centering
\linespread{1.3}\selectfont
\begin{tabular}{c|m{1.5cm}<{\centering}|m{1.5cm}<{\centering}|m{1.5cm}<{\centering}}
\toprule[1.5pt]

\multirow{2}{*}{\textbf{Method}} &
\multicolumn{3}{c}{\textbf{Metrics}} \\
\cline{2-4}
    & PSNR[dB] & SSIM[\%] & NMSE \\


\hline
3D VAE \cite{kingma2013auto} & 21.40 & 10.18 & 76.34 \\
\hline
3D VAE w/ WGAN-GP \cite{gulrajani2017improved} & 22.05 & 92.15 & 152.87 \\
\hline
PAVAE (Syn Mask) & 23.67 & 94.90 & 76.98 \\
\hline
PAVAE (Real Mask) & \textbf{32.74} & \textbf{99.29} & \textbf{15.68} \\

\bottomrule[1.5pt]
\end{tabular}
\end{table}

Quantitative results shown in Table \ref{tab:syn_results} indicate that neither 3D VAE nor 3D VAE w/ WGAN-GP can achieve high SSIM and low NMSE simultaneously. As for our model, we obtain relatively good synthetic results merely using generated masks. If we replace generated masks with real masks, the final results achieve the highest PSNR and SSIM, and lowest NMSE among all methods.

\subsubsection{Comparison of Lesion Segmentation Performance}

\begin{figure}[!t]
\centering
\includegraphics[width=0.95\textwidth]{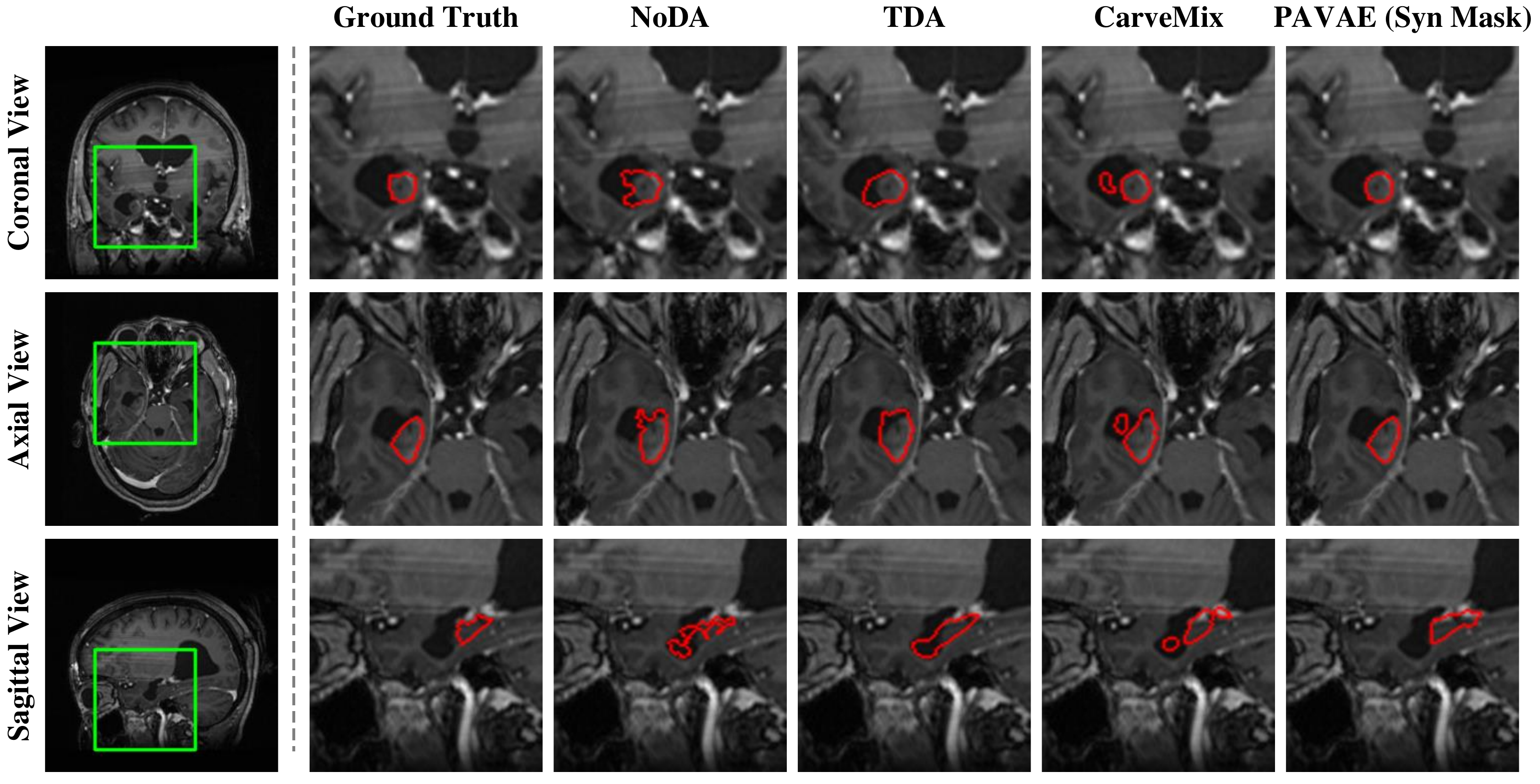}
\caption{Segmentation results from different training datasets for nnUNet models.} 
\label{fig:seg_viz}
\end{figure}

For the purpose of proving the effectiveness of synthetic results generated by our framework, we conduct the brain segmentation model based on nnUNet. For the training set and test set split, we follow the same manner with the lesion synthesis task. We generate 100 synthetic lesion images using CarveMix \cite{zhang2021carvemix} and PAVAE individually, and combine them with raw images to create new training datasets. We use Dice loss and Cross-entropy loss to train nnUNet for 200 epochs. Both quantitative and qualitative results are shown in Table \ref{tab:seg_results} and Fig. \ref{fig:seg_viz}. Here, NoDA means no data augmentation strategy is employed and only the real training dataset (37 samples) is used to train the model. TDA means traditional data augmentation strategies implemented in nnUNet, including random flip and rotations, and \etc. As shown in Table \ref{tab:seg_results}, our method has the best performance for three metrics (Dice, Jaccard, 95HD) and for the final metric ASD, only CarveMix has a slightly lower value. Furthermore, in Fig. \ref{fig:seg_viz}, we can observe that all the results of the competing methods over-segment the LITT ablation volume. Our method yields accurate segmentation results, which most closely resembles the expert annotation.

\begin{table}[!t]
\caption{Comparison of segmentation results using different data augmentation techniques during training.}
\label{tab:seg_results}
\centering
\linespread{1.3}\selectfont
\begin{tabular}{c|m{2cm}<{\centering}|m{2cm}<{\centering}|m{2cm}<{\centering}|m{2cm}<{\centering}}
\toprule[1.5pt]

\multirow{2}{*}{\textbf{Method}} &
\multicolumn{4}{c}{\textbf{Metrics}} \\
\cline{2-5}
    & Dice[\%] & Jaccard[\%] & ASD[voxel] & 95HD[voxel] \\


\hline
NoDA \cite{isensee2021nnu} & 66.69 & 51.19 & 1.17 & 3.52 \\
\hline
TDA \cite{isensee2021nnu} & 72.25 & 57.67 & 1.08 & 3.15 \\
\hline
CarveMix \cite{zhang2021carvemix} & 73.29 & 58.77 & \textbf{0.97} & 3.24 \\
\hline
PAVAE (Syn Mask) & \textbf{74.18} & \textbf{59.95} & 1.00 & \textbf{2.77} \\

\bottomrule[1.5pt]
\end{tabular}
\end{table}

\section{Discussion and Conclusion}
Building a 3D generative model may face several problems. The biggest ones can be mode collapses due to limited training samples and increasing computational complexity compared with 2D generative models. To tackle these issues, we have presented a progressive adversarial variational auto-encoder for brain lesion synthesis, which can generate reasonable masks and realistic brain lesions in a step-wise fashion. We further develop two types of blocks (\ie, CEB and MEB) to utilize both semantic and shape information to facilitate this lesion synthesis. Experimental results show that our framework can create high-fidelity brain lesions and boost the down-stream segmentation model training compared with existing methods. However, as can be found in the quantitative results (Fig. \ref{fig:syn_viz}), ground truth masks are still able to synthesize more realistic lesion images indicating a potential room for improvement when creating masks. Besides, all data used in this study was from a single center, further validation is required to evaluate its effectiveness on multi-center data and especially data acquired on different MRI scanners.

\subsubsection{Acknowledgement}

This work was supported by Centre for Doctoral Training in Surgical and Interventional Engineering at King's College London. This research was funded in whole, or in part, by the Wellcome Trust [218380/Z/19/Z, WT203148/Z/16/Z]. For the purpose of open access, the author has applied a CC BY public copyright licence to any Author Accepted Manuscript version arising from this submission. This research was supported by the UK Research and Innovation London Medical Imaging \& Artificial Intelligence Centre for Value Based Healthcare. The research was funded/supported by the National Institute for Health Research (NIHR) Biomedical Research Centre based at Guy's and St Thomas' NHS Foundation Trust and King's College London and supported by the NIHR Clinical Research Facility (CRF) at Guy's and St Thomas'. The views expressed are those of the author(s) and not necessarily those of the NHS, the NIHR or the Department of Health.


\bibliographystyle{splncs04}
\bibliography{books}

\end{document}